\documentclass[twocolumn]{article} 
\pdfoutput=1

\usepackage{amsmath}
\usepackage{amsfonts}
\usepackage{microtype}
\usepackage{graphicx}
\usepackage{booktabs}
\usepackage{algorithmic}
\usepackage{algorithm}
\usepackage[usenames,dvipsnames]{color}

\newcommand{\y}{\mathbf{y}}
\newcommand{\x}{\mathbf{x}}
\newcommand{\h}{\mathbf{h}}
\newcommand{\vu}{\mathbf{u}}
\newcommand{\va}{\mathbf{a}}
\newcommand{\tg}[1]{\textbf{#1}}
\newcommand{\given}{\,|\,}
\newcommand{\expectation}{\mathbb{E}}
\newcommand{\fe}{\mathcal{F}}
\DeclareMathOperator{\sigm}{sigm}

\newcommand{\twoheight}{1.9in}
\newcommand{\threeheight}{1.6in}

\newcommand{\crbmheight}{0.93in}

\newcommand{\citet}{\cite}

\usepackage[colorlinks=true]{hyperref}
\hypersetup{
  linkcolor = {RoyalBlue},
  citecolor = {RoyalBlue},
  urlcolor  = {RoyalBlue}
}

\title{Autotagging music with conditional restricted Boltzmann machines}

\author{ Michael Mandel$^1$\footnote{M.M. is currently at Audience, Inc.} \and Razvan Pascanu$^1$ \and Hugo Larochelle$^2$ \and Yoshua Bengio$^1$}

\date{Dept. IRO, Universit\'e de Montr\'eal$^1$ \hspace*{1cm} Comp.Sc. Dept. University of Toronto$^2$ }

\begin{document}

\maketitle

\begin{abstract}
  This paper describes two applications of conditional restricted
  Boltzmann machines (CRBMs) to the task of autotagging music.  The
  first consists of training a CRBM to predict tags that a user would
  apply to a clip of a song
  based on tags already applied by other users.  By learning the
  relationships between tags, this model is able to pre-process
  training data to significantly improve the
  performance of a support vector machine (SVM) autotagging.
  The second is the use of a discriminative RBM, a 
  type of CRBM, to autotag music.
  By simultaneously exploiting the 
  relationships among tags and between tags and audio-based features,
  this model is able to 
  significantly outperform SVMs, logistic regression, and multi-layer
  perceptrons.  In order to be applied to this
  problem, the discriminative RBM was generalized to the multi-label
  setting and four different learning algorithms for it were
  evaluated, the first such in-depth analysis of which we are aware.
\end{abstract}

\section{Introduction}

With the sizes of online music and media databases growing to millions
and billions of items, users need tools for searching and browsing
these items in intuitive ways.  One approach that has proven to be
popular is the use of social tags \cite{lamere08}, short descriptions
applied by users to items.  Users can search and browse through a
collection using the tags that they or others have applied.  This
system works well for popular items that have been tagged by many
users, but fails for items that are new or niche, this is the so-called cold
start problem \cite{ScheinEtAl2002}.

One promising way to overcome the cold start is through content-based
analysis and tagging of the items in the collection, known as
autotagging.  Researchers have investigated a number of autotagging
techniques for music over the last decade \cite{whitman02, eck08,
  tingle10}.  While a few autotagging techniques attempt to capture
the relationship between tags (e.g.\ \cite{bertinmahieux08}), many
treat each tag as a separate classification or ranking problem (e.g.\
\cite{mandel08b}).  The problem of predicting the presence or
relevance of multiple tags simultaneously is known as the
multi-label classification problem \cite{trohidis+etal:2008}.

This paper explores techniques for autotagging music that incorporate
the relationships between tags.  We approach this problem in two ways,
both of which are based on conditional restricted Boltzmann machines (RBMs)
described in Section~\ref{sec:rbms}.  The first approach, described in
Section~\ref{sec:smoothing}, is a novel model 
trained to predict the tags that a user will apply to music based on
the tags other users have applied to it.  It is a purely textual model
in that it does not utilize the audio at all to make 
predictions.  These predicted tags, which we call ``smoothed'' tags,
are then used to train different types of classifiers that do utilize
audio.

The second approach, described in
Section~\ref{sec:drbm}, is a discriminative RBM
\cite{Larochelle+Bengio-2008}, which learns to jointly predict tags
from features extracted from the audio.  We extend the discriminative
RBM to perform multi-label classification instead of the
winner-take-all classification performed by previous discriminative
RBMs.  This new model requires a new training algorithm.  We explore
four techniques for approximating the gradient of the model parameters, namely maximum
likelihood using contrastive divergence, maximum 
pseudo-likelihood, 
mean-field contrastive divergence, and loopy belief propagation
approximations.

Section~\ref{sec:experiments} investigates the performance of these
two methods separately and together on three different datasets, two
of which have been previously described in the literature, and one of
which (the largest of the three) is new and has not been used to train or test
autotaggers before.

\section{Restricted Boltzmann machines} \label{sec:rbms}

This section describes the restricted Boltzmann machine (RBM)
\cite{smolensky86}, its conditional variant the conditional RBM, and
one particular type of conditional RBM, the discriminative RBM.  
 The RBM is an
undirected graphical model that generatively models a set of input
variables $\y=(y_1,\ldots,y_C)^T$ with a set of hidden variables
$\h=(h_1,\ldots,h_n)^T$.  Both $\y$ and $\h$ are typically binary,
although other distributions are possible.  The model is
``restricted'' in that the dependency between the hidden and visible
variables is bipartite, meaning that the hidden variables are
independent when conditioned on the visible variables and vice versa.
The joint probability density function is
\begin{equation}
p(\y,\h) = \frac{1}{Z} e^{-E(\y,\h)}
\end{equation}
where
\begin{equation}
E(\y,\h) = -\h^T U \y - c^T \h - d^T \y,
\end{equation}
\begin{equation}
Z = \sum_{\y,\h} e^{-E(\y,\h)}, \label{eq:partition}
\end{equation}
$U$ is a matrix of real numbers, and $c$ and $d$ are vectors of real
numbers.
The computation of $Z$, known as the partition function, is
intractable due to the number of terms being exponential in 
the number of units.  
The marginal of $y$, however, is $p(\y) = e^{-\fe(\y)}/Z$, where 
$\fe(\y)$ is the free energy of $\y$ and can be easily 
computed as 
\begin{equation}
\fe(\y) = - \log \sum_\h e^{-E(\y,\h)} = -d^T \y - \sum_i \log
(1+e^{c_i + U_i \y}).
\end{equation}
The parameters
of the model can be optimized using gradient descent to minimize the
negative log likelihood of data $\{\y_t\}$ under this model
\begin{equation}
  \frac{\partial}{\partial \theta} p(\y_t) = -\expectation_{\h
    \given \y_t} \left[ \frac{\partial}{\partial \theta}
    E(\y_t,\h) \right] + \expectation_{\y, \h} \left[
    \frac{\partial}{\partial \theta} E(\y,\h)
  \right]. \label{eq:joint_grad}
\end{equation}
The first expectation in this expression is easy to compute, but the
second is intractable and must be approximated.  One popular
approximation for it is contrastive divergence \cite{hinton02}, which
uses a small number of Gibbs sampling steps \emph{starting from the
  observed example} to sample from $p(\y,\h)$.

\begin{figure*}[t]
\centering
\makebox[\textwidth]{%
\begin{tabular}{ccc}
\includegraphics[height=\crbmheight]{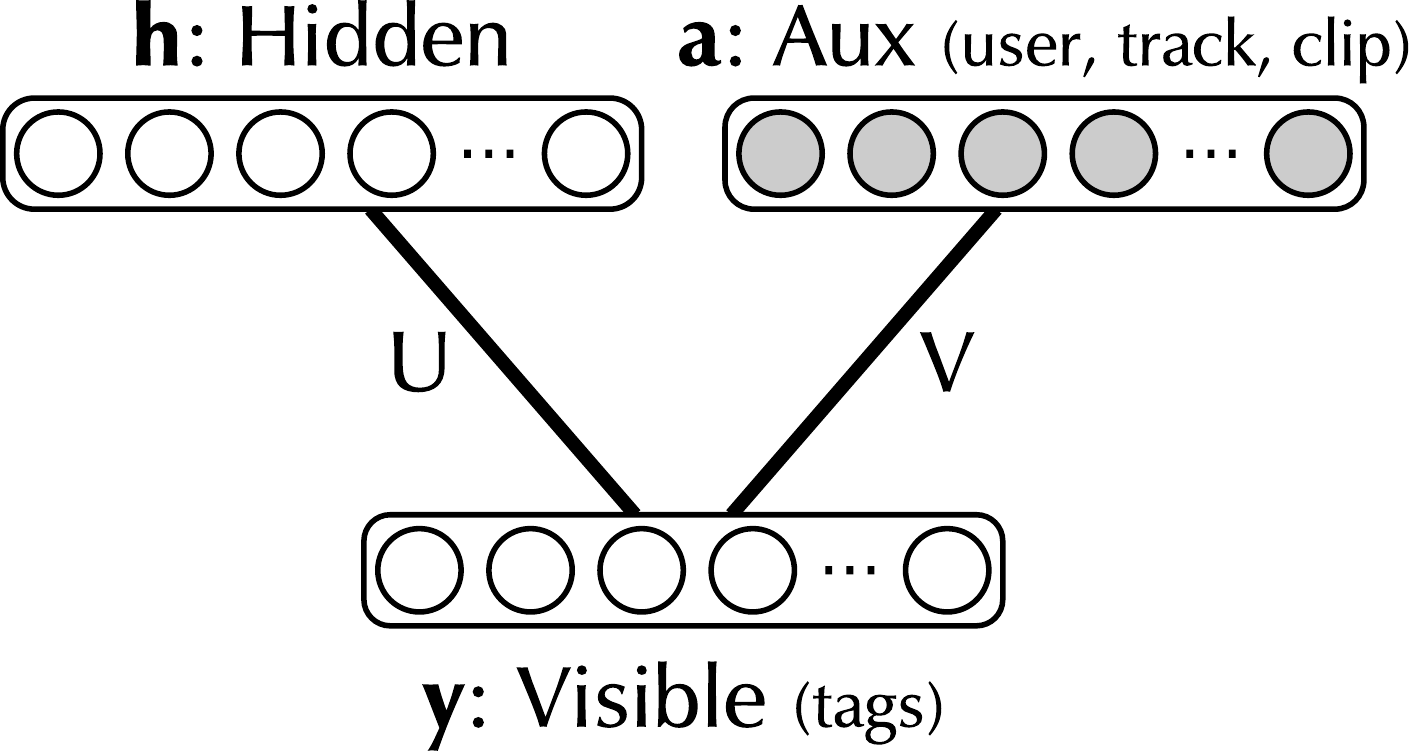} &
\includegraphics[height=\crbmheight]{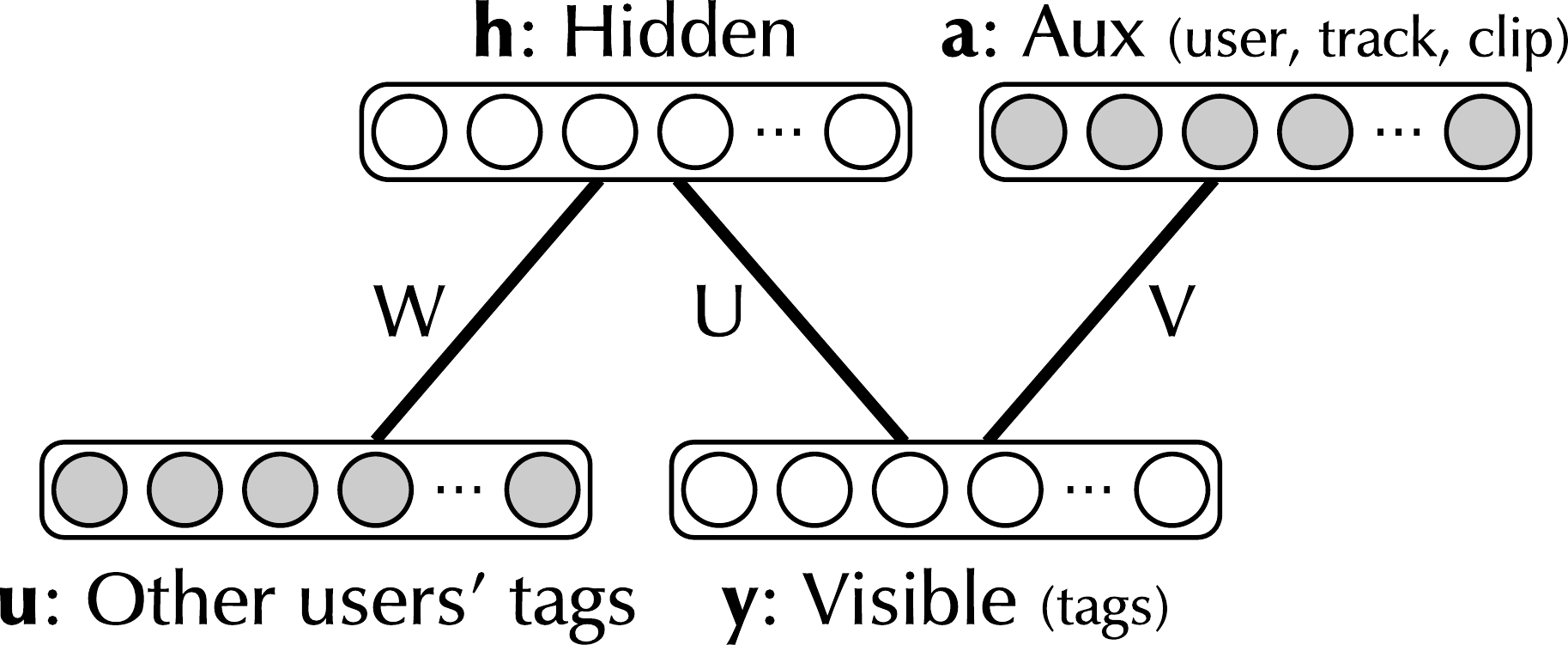} &
\includegraphics[height=\crbmheight]{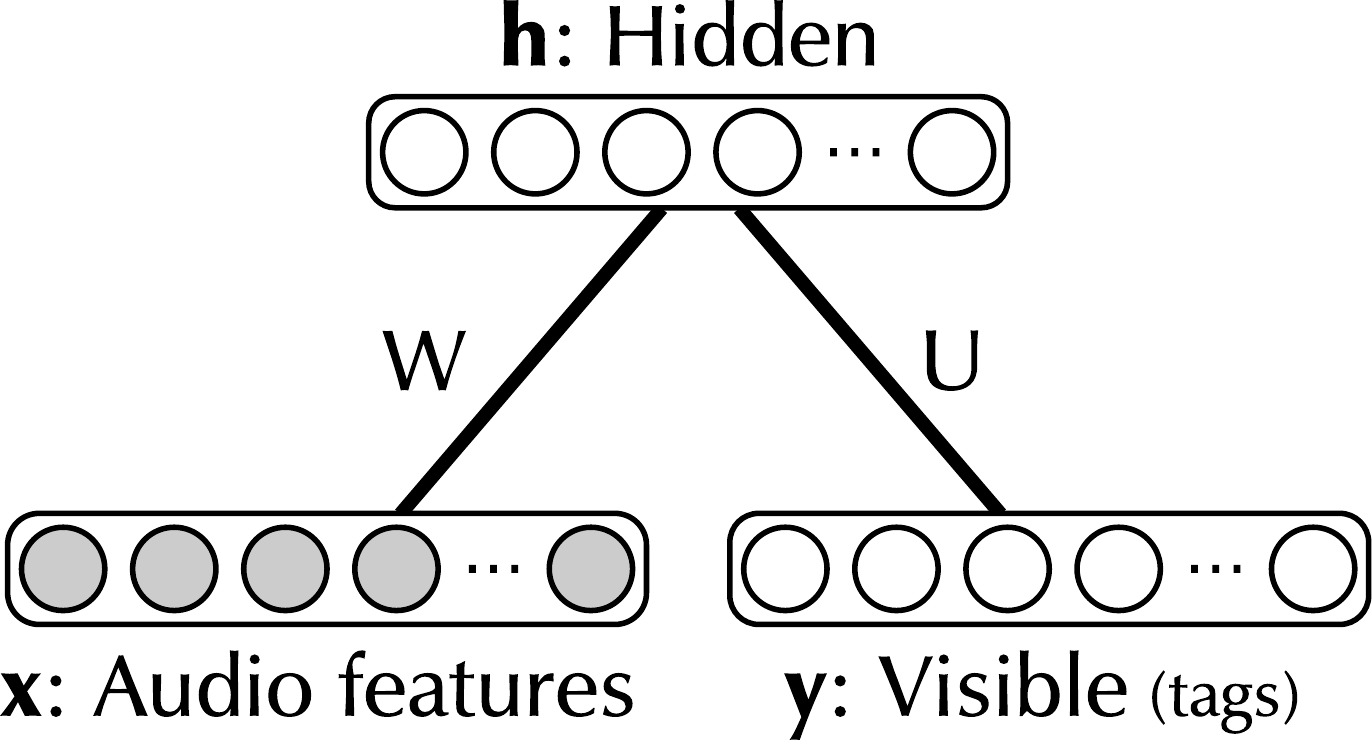} \\
(a) & (b) & (c) \\
\end{tabular}}
\caption{Schematic diagrams of the various restricted Boltzmann
  machines under investigation. (a) RBM for tag smoothing conditioned
  on just auxiliary information: user, track, clips identity, (b) RBM
  for tag smoothing conditioned on auxiliary information and tags of
  other users, (c) discriminative RBM for audio classification.
  Filled circles show variables that are always observed, open circles
  show variables that are inferred at test
  time.} \label{fig:crbms}
\end{figure*}

RBMs can be conditioned on other variables~\cite{Taylor+2007}.  In
general, as shown in Figure~\ref{fig:crbms}(b), both the hidden
and visible units can be conditioned on other variables,
$\vu=(v_1,\ldots,v_d)^T$ and $\va=(a_1,\ldots,a_A)^T$, respectively.
Including these interactions, the energy function becomes
\begin{equation}
E(\y,\h,\vu,\va) = -\h^T U \y - \h^T W \vu - \y^T V \va - d^T \y - c^T \h \label{eqn:energy}
\end{equation}
and $p(\y,\h \given \vu, \va) \propto e^{-E(\y,\h,\vu,\va)}$, where
$W$ and $V$ are real matrices.  The 
vectors $V \va$ and $W \vu$ act like additional biases on $\y$ and
$\h$.  By setting the appropriate $W$ or $V$ matrix or conditioning
vector $\vu$ or $\va$ to 0, the conditioning
can apply to only the visible units, as in
Figure~\ref{fig:crbms}(a), or only the hidden units, as in
Figure~\ref{fig:crbms}(c).  For an observed data point
$\y_t,\vu_t,\va_t$, the gradient of the log likelihood 
with respect to a parameter $\theta$ becomes
\begin{multline}
  \frac{\partial}{\partial \theta}\log p(\y_t \given \vu_t, \va_t) = 
  -\expectation_{\h \given \y_t, \vu_t, \va_t} \left[
    \frac{\partial}{\partial \theta} E(\y_t,\h,\vu_t,\va_t) \right] \\
  + \expectation_{\y, \h \given \vu_t, \va_t}
  \left[\frac{\partial}{\partial \theta} E(\y,\h,\vu_t,\va_t)
  \right]. \label{eq:cond_grad}
\end{multline}
\cite{salakhutdinov07} describes a conditional RBM used for
collaborative filtering in which only the hidden variables are
conditioned on other variables.

\subsection{Conditional RBMs for tag smoothing} \label{sec:smoothing}

We first employ conditional RBMs to learn relationships
among tags and between tags and users, tracks, and audio clips of tracks.
This model is purely textual, meaning that it only operates on the
tags and not on the audio.

All of the datasets used in this paper were collected by open endedly
soliciting tags from users to describe audio clips.  This means that the
tags that they contain are most 
likely relevant, but the tags that  are not present are not
necessarily irrelevant.  Thus there is a need to distinguish tags
omitted but still relevant from those that do not apply, as well as
tags that were included erroneously from those that truly apply.  As
shown in \cite{mandel10b}, 
the co-occurrences of tags can be used to predict both of these cases.
For example, if the tags \tg{rap} and \tg{hip-hop} 
frequently co-occur and a clip has been tagged \tg{hip-hop} but not
\tg{rap}, it would be reasonable to increase the likelihood of
\tg{rap} being relevant to that clip, although perhaps not as much as
if it had actually been applied by a user.  Similarly, it might be
reasonable to decrease the likelihood of \tg{hip-hop} being relevant
as it was not corroborated by an application of \tg{rap}.

We use the {\em doubly conditional RBM} shown in Figure~\ref{fig:crbms}(b)
for this sort of tag ``smoothing'' as we call it.
The binary visible units
represent the tags that one user has applied to a clip and the hidden
units capture second order relationships between these tags.  The
visible units are conditioned on auxiliary variables $\va$ which
represent as one-hot vectors the user, track, and clip from which a
vector of tags is observed.  The hidden units are conditioned on
auxiliary variables $\vu$, which represent the tags that other users
have applied to the same clip.  

The vectors $\y$ and $\vu$ are the same size, but whereas $\y$ is a
binary vector representing which of the fixed vocabulary of tags the
target user applied to the target clip, $\vu$ is a vector of the
average of these binary vectors for all of the other users who have
seen the target clip.  Thus the values in $\vu$ are still between 0
and 1, but are continuous-valued.  At test time, $\vu$ is set to the
average tag vector of all of the users and predicts the tags that a
new user would likely apply to the clip given the tags that other
users have already applied.

The weights $V$ and $W$ are penalized with an $\ell_1$ cost to
encourage them to only capture dependencies that depend on specific
settings of the auxiliary variables and push into $U$ the dependencies
that exist independently of the auxiliary variables.  This means that
$V$ should ideally only capture tag information relevant to a
particular user, clip, or track, $W$ should capture information about
the relationships between other user's tags and the current user's
tags, and $U$ should capture information about the co-occurrences of
tags in general.

Compare this to the singly
conditional RBM shown in Figure~\ref{fig:crbms}(a) and
described in \cite{mandel10b}.  This CRBM also includes the conditioning of
the visible units on 
the user, clip, and track information, but does not include the
conditioning on other users' tags.  While the doubly conditional RBM
can use its modeling power to learn to predict specific user's tags
from general tag patterns, this singly conditional model must predict
both the general tag patterns and specific user's tags, a harder
problem.  We found that the doubly-conditional RBM's smoothing trains
better SVMs on a validation experiment, and so we did not include the
singly-conditional RBM in our experiments.

\section{Discriminative RBMs} \label{sec:drbm}

One useful variant of the conditional RBM is the discriminative RBM
\cite{Larochelle+Bengio-2008}, shown in Figure~\ref{fig:crbms}(c).  
The discriminative RBM is a conditional RBM that is trained to
predict the probability of the class labels, $\y$, from the rest of
the inputs, $\x$. Based on the energy function of \eqref{eqn:energy},
it corresponds to setting $\vu = \x$ and $\va=0$.


For a set of observed data points $\{(\y_t,\x_t)\}$, the
discriminative RBM optimizes the log
conditional, $\log p(\y_t \given \x_t)$, i.e.\ focusing on predicting
$\y_t$ from $\x_t$ well. A generative variant of this RBM would
instead optimize $\log p(\y_t, \x_t)$, the joint distribution 
(in this case, $\x_t$ acts
as an extension of $\y_t$, i.e.\ it is not conditioned on it).

Looking at the parameter gradient of \eqref{eq:cond_grad},
we see that the second expectation requires a sum
over all configurations of $\y$.  When $\y$ can take only a
few values, as in ordinary classification
tasks~\cite{Larochelle+Bengio-2008}, this expectation can be computed
efficiently and exactly. However, here $\y$ is a set of $C$ binary
indicators (the presence of a tag) that are not mutually exclusive, so
that the expectation has $2^C$ terms and must be approximated because
it cannot be computed in closed form.
Note that given a value for $\y$, $p(\h|\y,\x_t)$
factors and is computed exactly.

\subsection{Approximations to the expectation} \label{sec:approx}

In the case of the discriminative RBM involving $\y$, $\h$, $\vu =
\x$, and $\va = 0$, we approximate the $\expectation_{\y,\h \given
  \x_t}$ term in \eqref{eq:cond_grad} in three different ways: using
contrastive divergence, mean-field contrastive divergence, and loopy
belief propagation.  We also compare a similar, but tractable
computation that maximizes the pseudo-likelihood.  The difficulty in
computing this expectation stems directly from the difficulty in
computing $p(\y,\h \given \x_t)$, which in turn is caused by the
interdependence of the $\y$ and $\h$ variables.

Contrastive divergence (CD) \cite{hinton02} has proven to be a very
popular algorithm for estimating the log-likelihood gradient in RBMs,
and it can also be used in the case of conditional RBMs.
Typically, it is used to compute
$\expectation_{\y,\x,\h}[\cdot]$ as opposed to here, where we compute
$\expectation_{\y,\h \given \x_t}[\cdot]$.  To compute the usual
CD-$k$ update, $k$ steps of block Gibbs sampling, {\em starting
from the observed example} $(\x_t,\y_t)$, are used to
approximate the expectation. The block Gibbs chain is obtained 
by alternating sampling from $p(\h \given
\y, \x)$ and sampling from $p(\y,\x \given \h)$.  In the case of the
conditional CD, we sample from $p(\h \given \y, \x_t)$ and then from
$p(\y \given \h)$ (since $\h$ isolates $\y$ from $\x_t$), 
keeping $\x_t$ fixed throughout.  CD can be noisy
because it uses a small number of samples (usually only one), and it can be biased
because it doesn't necessarily run the Markov chain to convergence (usually only
1 to 10 steps).

The mean-field contrastive divergence approach approximates the $\y$
and $\h$ variables using 
their conditional expectations (given each other)
and iteratively updates each one based on the
estimate of the other until convergence (note that $\x_t$ is fixed).
\begin{align}
\h^k &= \expectation[\h \given \y^{k-1}, \x_t] = \sigm(c +
U \y^{k-1} + W \x_t) \\
\y^{k} &= \expectation[\y \given \h^{k}, \x_t] = \sigm(b +
W \h^{k}).
\end{align}
In this case, we plug the continuous-valued expectations into these
equations instead of the sampled binary values that should formally be
used.  While this method is
straightforward, it cannot capture multimodal distributions in $\y$
and $\h$, which makes it sensitive to initialization.  We set the
initial condition $\y^{-1}=\y_t$, i.e.\ we initialize $\y$ at the
training label from which we compute $\h^0$, etc., which is why this is
referred to as mean-field contrastive divergence.  We also tried to
use standard mean-field where $\y^{-1}=0$, but found the results to be
much worse.

Loopy belief propagation~\cite{Murphy99loopybelief} (LBP) is another
algorithm for approximating intractable marginals in a graphical
model. It relies on a message passing procedure between the variables
of the graph. While not guaranteed to converge it frequently does in
practice, and gives estimates of the true marginals that are often
more accurate than the iterative mean-field
procedure~\cite{WeissY2001}.  In this setting, we used LBP to
estimate the marginals $p(y_j=1|\x_t)$, $p(h_k=1|\x_t)$ and 
$p(y_j=1,h_k=1|\x_t)$ for a given $\x_t$ under the discriminative RBM,
and used those marginals to compute the $\expectation_{\y,\h \given
  \x_t}$ term in equation \eqref{eq:cond_grad}. The quantity
$p(y_j=1|\x_t)$ can also be estimated at test time to predict the
labels. One method that has been shown to be useful in aiding
convergence is message damped belief propagation \cite{pretti05}.  In
this case the updates computed by belief propagation are mixed
with the previous updates for the same variables in order to smooth
them, the damping factor being a parameter of the algorithm.

Another method for tuning the parameters aims to optimize not the
likelihood of the data, but the pseudo-likelihood \cite{besag75}.  The
pseudo-likelihood circumvents the intractability of 
computing the partition function in \eqref{eq:partition} by
considering only configurations of the visible units that are within a
Hamming distance of 1 from the training observation.
\begin{align}
\log PL(\y \given \x) &= \sum_j \log p(y_j \given
\y_{\backslash j}, \x) \\
= \sum_j &\log p(\y \given \x) - \log \left(
  p(\y \given \x) + p(\tilde{\y}_j \given \x) \right) \nonumber
\end{align}
where $\y_{\backslash j}$ is the labels vector $\y$ without the $j$th
variable and $\tilde{\y}_j$ is the labels vector $\y$ with the $j$th bit
flipped (the subscript $t$ is removed here for clarity). The pseudo-likelihood can be optimized using gradient descent.

Because of lack of space, we give pseudocodes for all the
aforementioned algorithms 
in the appendix. Additionally, the
python code used for training these models is available on our
website\footnote{\url{http://www.iro.umontreal.ca/~bengioy/code/drbm_tags}}.
Note that while all of these methods can be used for training, not all
of them can be used at test time to estimate $P(y_j=1|\x_t)$.  
Specifically, the pseudo-likelihood
requires the knowledge of $\y_{\backslash i}$, which is unavailable at
test time.  Similarly, CD must be initialized from the true values of
$\y_t$ and $\x_t$.  It is possible to use a Gibbs sampling method
similar to CD starting from an arbitrary initialization of $\y_t$, but
this is costly because the Markov chain may need to be run for many
iterations before it mixes well.  We found that mean-field CD could be
successfully initialized with $\y^{-1}=0$ at test time.

\section{Experiments} \label{sec:experiments}

We performed a number of experiments to compare 
different hyper-parameter settings, to compare different classifiers, and to
compare different tag smoothing techniques.
These experiments were based on three different datasets: data from
Amazon.com's Mechanical Turk service\footnote{\url{http://mturk.com}},
data from the MajorMiner music labeling
game\footnote{\url{http://majorminer.org}}, and data from Last.fm's
users\footnote{\url{http://last.fm}}.  We compare the 
discriminative RBM to standard (generative) RBMs, multi-layer
perceptrons, logistic 
regression, and support vector machines.  All of these algorithms were
evaluated in terms of retrieval performance using the area under the
ROC curve.

\subsection{Datasets}
Three datasets were used in these experiments.  All of these datasets
were in the form of (user, item, tag) triples, where the items were
either 10-second clips of tracks or whole tracks.  These data were
condensed into (item, tag, count) triples by summing across users.

The first dataset was collected from Amazon.com's Mechanical Turk service and is
described in \cite{mandel10b}.  Users were asked to describe 10-second
clips of songs in terms of 5 broad categories including genre,
emotion, instruments, and overall production.  The music used in the
experiment consisted of 185 songs selected randomly from the music
blogs indexed by the Hype Machine\footnote{\url{http://hypem.com}}.
From each track, five 10-second clips were extracted from
proportionally equally spaced points, for a total of 925 clips.  Each
clip was seen by a total of 3 users, generating approximately 15,500
(user, clip, tag) triples from 210 unique users. We used the most
popular 77 tags for this dataset.

The second dataset was collected from the MajorMiner music labeling
game and is described in \cite{mandel08b}.  Players were asked to
describe 10-second clips of songs and were rewarded for agreeing with
other players and for being original.  This dataset includes
approximately 80,000 (user, clip, tag) triples with 2600 unique clips,
650 unique users, and 1000 unique tags. We used the most popular
77 tags for this dataset.

The final dataset was collected from Last.fm's website and is
described in \cite{SchifanellaEtAl2010}.  The entire dataset consists of
approximately 7 million (user, track, tag) triples from 84,000 unique
users, 1 million unique tracks, and 280,000 unique tags.  While only
the textual information was collected from Last.fm, we were able to
match it to 47,000 tracks in our own music collection.  While this may
seem like a small fraction of the total number of tracks, the tracks
that were found included 1.5 million of the (user, track, tag)
triples, implying that the tracks we were able to match were tagged
more often than average.  Following similar reasoning, many of these
users, tracks, and tags occurred infrequently, with 1
million (user, track, tag) triples in which all three items occurred in
at least 25 triples.  Because these tags were
applied at the track level and not at the clip level, we selected 
one clip from the center of
each track and assumed that they
should all be described with the track tags.  This is the simplest
solution to this problem, although using some form of
multiple-instance learning might find a better solution
\cite{mandel08a}. We used the most popular 100 tags for this dataset.

Converting (item, tag, count) triples to binary matrices for training
and evaluation purposes required some care.  In the MajorMiner and
Last.fm data, the counts were high enough that we could require the
verification of an (item, tag) pair by at least two people, meaning
that the count had to be at least 2 to be considered as a positive
example.  The Mechanical Turk dataset did not have high enough counts
to allow this, so we had to count every (item, tag) pair.  In the
MajorMiner and Last.fm datasets, (item, tag) pairs with only a single
count were not used as negative examples because we assumed that they
had higher potential relevance than (item, tag) pairs that never
occurred, which served as stronger negative examples.

\paragraph{Features}
The timbral and rhythmic features of \cite{mandel08b} were used to
characterize the audio of 10-second song clips.  The timbral features
were the mean and rasterized 
full covariance of the clip's mel frequency cepstral coefficients.
They capture information about instrumentation and overall production
qualities.  The rhythmic features are based on the modulation spectra
in four large frequency bands.  In fact, they are closely related to
the autocorrelation in those frequency bands.  They capture
information about the rhythm of the various parts of the drum kit (if
present), i.e.\ bass drum, tom tom, snare, hi-hat.  They also
discriminate between music that has a strong rhythmic component, e.g.\
dance music, and music that does not, e.g. folk rock.  Each dimension
of both sets of features was normalized across the database to have
zero-mean and unit-variance, and then each feature vector was normalized
to be unit norm to reduce the effect of outliers.  The timbral
features were 189-dimensional and the rhythmic features were
200-dimensional, making the combined feature vector 389-dimensional.

\subsection{Classifiers}

We compared a number of classifiers including two variants of
restricted Boltzmann machines, and three other standard classifiers.
The RBMs we compared were the discriminative RBM, described in
Section~\ref{sec:drbm}
and a standard generative RBM. Both RBMs 
use Gaussian input units~\cite{welling05} in order to deal with the
continuous-valued features for $\x$. The 
other classifiers include a multi-layer perceptron, logistic
regression, and support vector machines. For all datasets we select 
the hyper-parameters of the model using a 5-fold cross-validation. In order
to increase accuracy of our measure, 
for each fold we computed the score as an average across 4 sub-folds. Each 
run used a different fold (from the remaining 4 folds) as the validation set 
and the other 3 as the training set.

The discriminative RBM uses the gradient updates shown in
\eqref{eq:cond_grad}, while the generative RBM uses a different update
in which the second expectation is $\expectation_{\y,\x,\h}$ instead
of $\expectation_{\y,\h \given \x_t}$.  The generative RBM attempts to
maximize $\log p(\y,\x)$, while the discriminative RBM attempts to maximize
$\log p(\y \given \x)$.  It is also possible to use a mixture of these
two objective functions and maximize $\alpha \log p(\y,\x) + \log p(\y
\given \x)$, referred to as a hybrid
generative/discriminative RBM~\cite{Larochelle+Bengio-2008}.  In our
experiments, however, the hybrid RBM did
not improve on the DRBM, so we will not discuss it further.  For each
model and dataset pair we 
optimized the hyper-parameters using the cross validation described
above, selecting the hyper-parameters with the best performance on the
validation set averaging across folds and tags. Different
hyper-parameters performed best in each case, which is to be expected given the 
differences in the models and in the data.  For example, one would expect the
generative RBM to require more hidden units than the discriminative RBM because
it models the joint probability.  Also on a
large dataset, one would expect to be able to use more hidden units without
overfitting. The hyper-parameters that performed best on the validation set
can be seen in Table~\ref{tab:params}.

\begin{table}[t]
\caption{Parameter settings found to perform best on validation
  sets and used in experiments for discriminative and generative RBMs,
  multi-layer perceptrons, and logistic 
  regression. LR stands for learning rate.} \label{tab:params}
\begin{center}
\begin{tabular}{lcccc}
\toprule
 & & \multicolumn{3}{c}{Number of hidden units} \\
Model       & LR & MTurk & MajMin & Last.fm \\
\midrule
Disc.\ RBM  & 0.01  & 50  & 100 & 200 \\
Gen.\ RBM   & 0.01  & 200 & 300 & 300 \\
MLP         & 0.001 & 250 & 250 & 250 \\
Log.\ reg.\ & 2.0   & --- & --- & --- \\
\bottomrule
\end{tabular}
\end{center}
\end{table}

The multi-layer perceptron (MLP) is quite similar in structure to the
discriminative RBM in that it has nodes representing the features and
the classes and hidden nodes that capture interactions between them.  The
main difference is that in estimating $p(\y \given \x)$ there is no modeling of the
interactions between the elements of $\y$ given $\x$.  In the
discriminative RBM, at test time the unknown $\y$ and $\h$ interact
with one another through one of the methods described in
Section~\ref{sec:approx} until they reach a mutually agreeable
equilibrium.  In the case of the MLP, however, at test time $\h$ is
computed deterministically from $\x$ and $\y$ is computed
deterministically from $\h$.  The stochastic hidden units in the
discriminative RBM at test time allow it to better capture
interactions between the variables in $\y$ (i.e. the tags).

An even simpler classifier than the MLP is logistic regression, which
has no hidden layer and predicts each class directly from the input
features.  We similarly optimize this using gradient descent, where
the cost function is the cross-entropy between the target
labels and the predictions, like for the MLP.  



The final classifier we compared is the support vector machine (SVM).
Specifically we used a linear kernel and a $\nu$-SVM
\cite{scholkopf00} to automatically select 
the $C$ parameter.  We trained a different SVM for each tag as
an independent two-way decision (e.g. \tg{rock} vs not \tg{rock}).
While the above methods based on stochastic 
gradient descent can be trained on all examples, SVMs are more
sensitive to the relative number of positive and negative examples, so
we had to more carefully select the training examples to use for each
tag.  To do this, we selected as positive examples those clips to which
users applied a given tag most frequently and as negative examples
those clips to which users applied a given tag least frequently
(generally 0 times).  The actual training labels used, however, were
still the standard $\pm1$ targets.  We ensured that there were the same number
of positive and negative examples, up to 200 of each.

\paragraph{Metrics}
The performance of all of these algorithms on all of these datasets is
evaluated in terms of retrieval performance using the area under the
ROC curve (AUC) \cite{CortesAndMohri2004}.  This metric scores the
ability of an algorithm to rank relevant examples in a collection above
irrelevant examples.  A random
ranking will achieve an AUC of approximately 0.5, while a perfect
ranking will achieve an AUC of 1.0.  
In certain experiments CRBMs were used to smooth the training data,
but the testing data was always the unsmoothed, user-supplied tags.
We measure the AUC for each tag separately.
We use the average across tags and folds as a overall measure of performance 
and consider the standard error across folds for Figure~\ref{fig:params}. 
For a more detailed comparison we use a two-sided paired t-test across folds,
per tag, between two models.  We count the number
of tags for which each model performs better than the other at a 95\% significance level.

\paragraph{Implementation details / Running time} 

In order to find the parameters that worked best for the DRBM, we used
a grid search.  To avoid a prohibitive number of combinations, we
settled on a learning rate and number of hidden units before exploring
gradient approximations, Loopy Belief Propagation damping factors, and
numbers of iterations for CD, MF-CD or LBP. We also performed a much
wider parameter search on the smaller datasets, MTurk and MajorMiner,
keeping the same parameters for Last.fm, but varying the number of
hidden units. We found that the DRBM is insensitive to the number of
iteration steps while the computational cost increases considerably.
Training time varies according to many details, but on average, to
train a DRBM on MajorMiner took around 48 CPU-hours.

\subsection{Experiments}

\begin{figure}[t]
\begin{center}
\includegraphics[height=\twoheight]{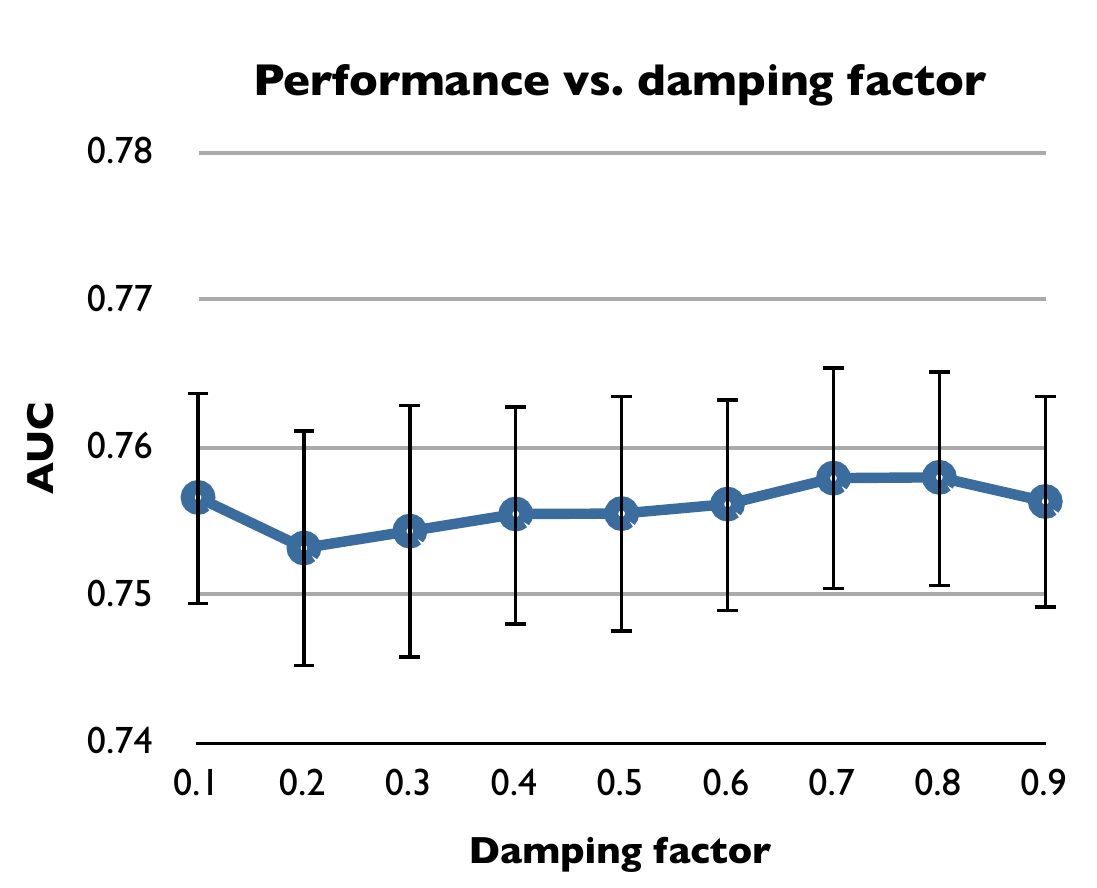} 
\end{center}
\caption{ Average area under the ROC curve with 
standard errors on the MajorMiner dataset for the discriminative RBM trained
using loopy belief propagation with different damping factors.}
\label{fig:params1}
\end{figure}

\begin{figure*}[t]
\begin{center}
\makebox[\textwidth]{%
\begin{tabular}{cc}
\includegraphics[height=\twoheight]{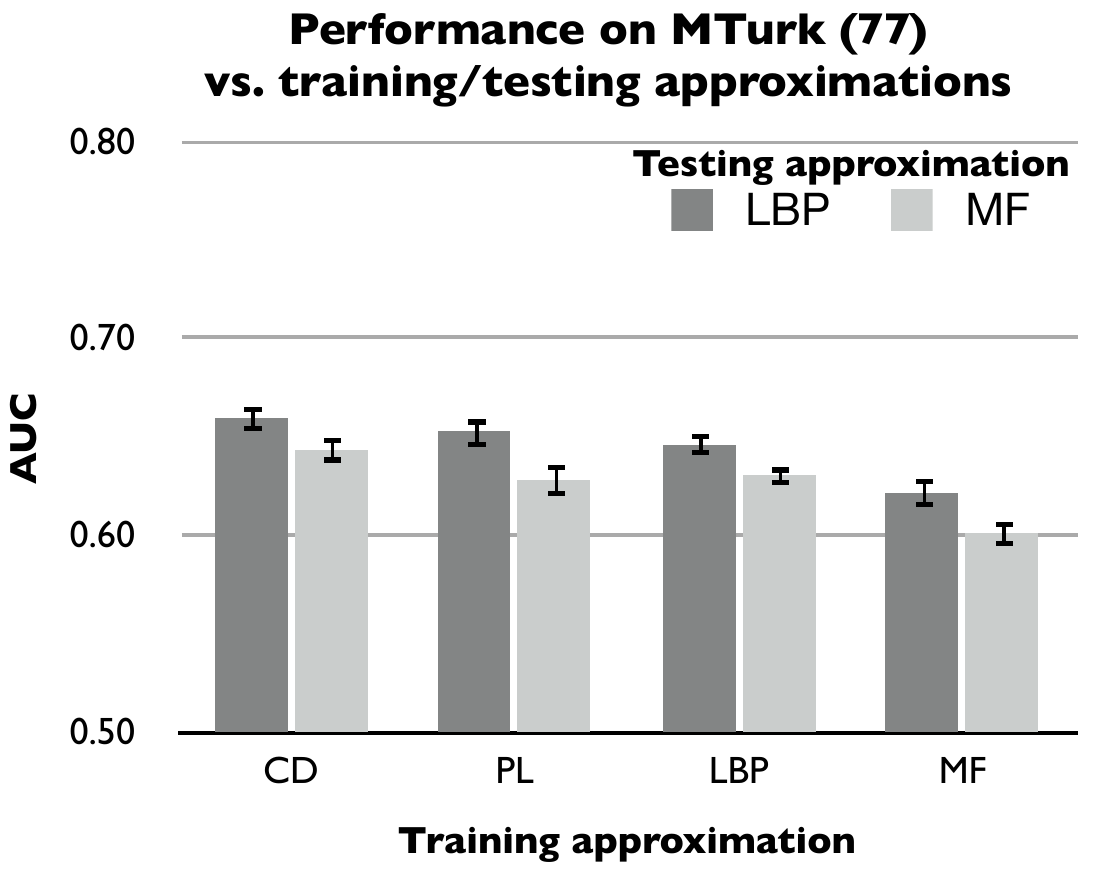} &
\includegraphics[height=\twoheight]{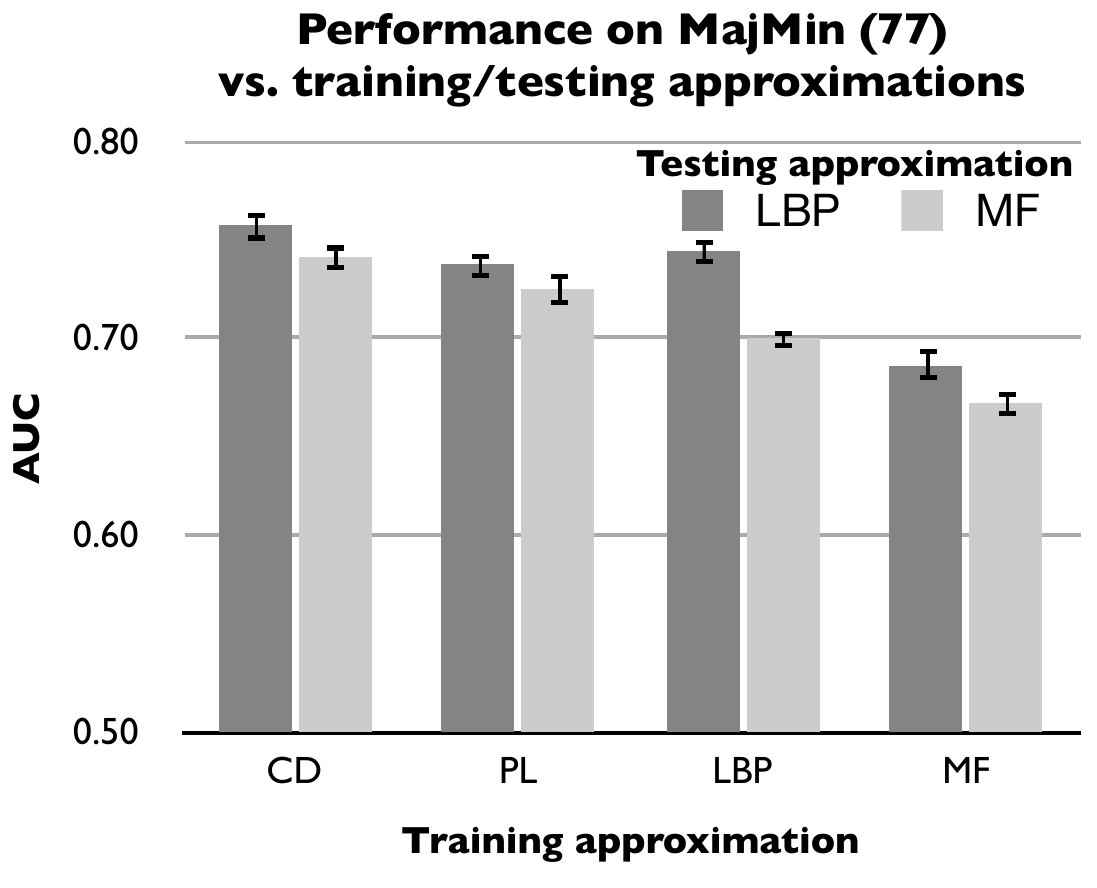} \\
\end{tabular}}
\end{center}
\caption{Results on Mechanical Turk and MajorMiner comparing the
performance of different approximations 
for the discriminative RBM during training and testing: contrastive
divergence (CD),  
pseudo-likelihood (PL), mean field contrastive divergence (MF) and loopy 
belief propagation (LBP). The approximations used during training are represented
on the x-axis, while the approximations used during testing are represented through
the gray value of the bar.}
\label{fig:params}
\end{figure*}

\begin{figure*}[t]
\begin{center}
\makebox[\textwidth]{%
\begin{tabular}{cc}
\includegraphics[height=\threeheight]{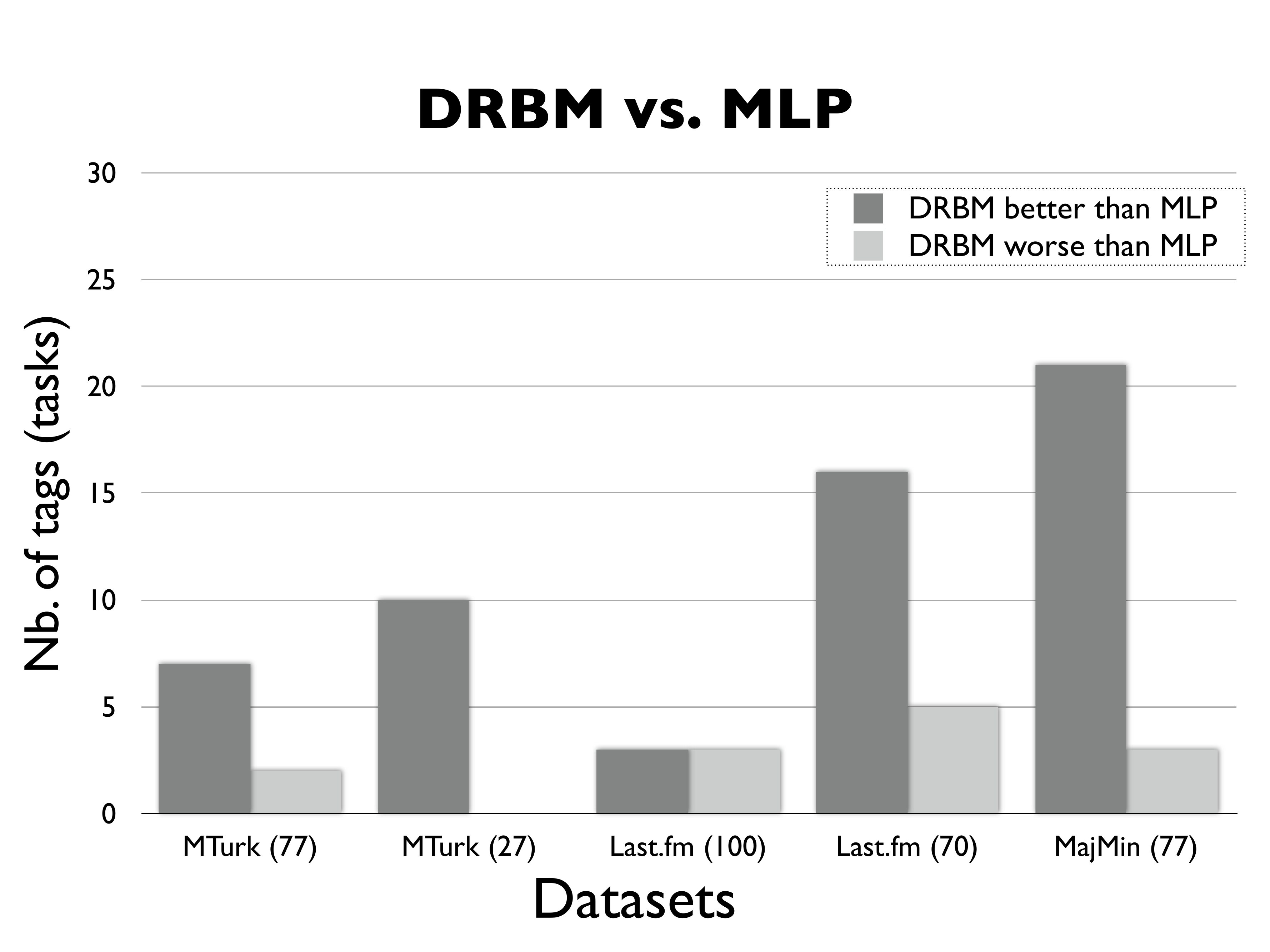} &
\includegraphics[height=\threeheight]{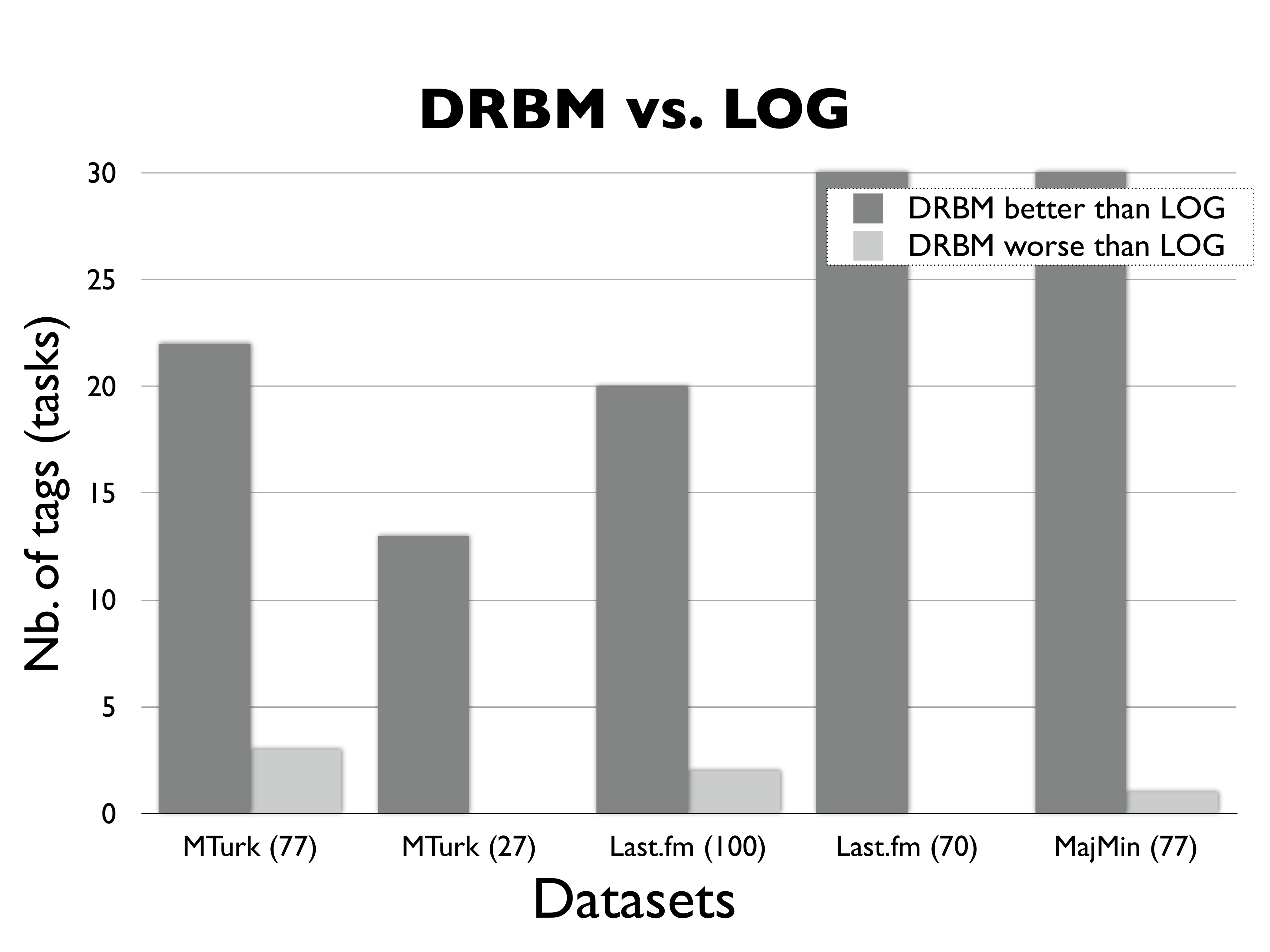} \\
\includegraphics[height=\threeheight]{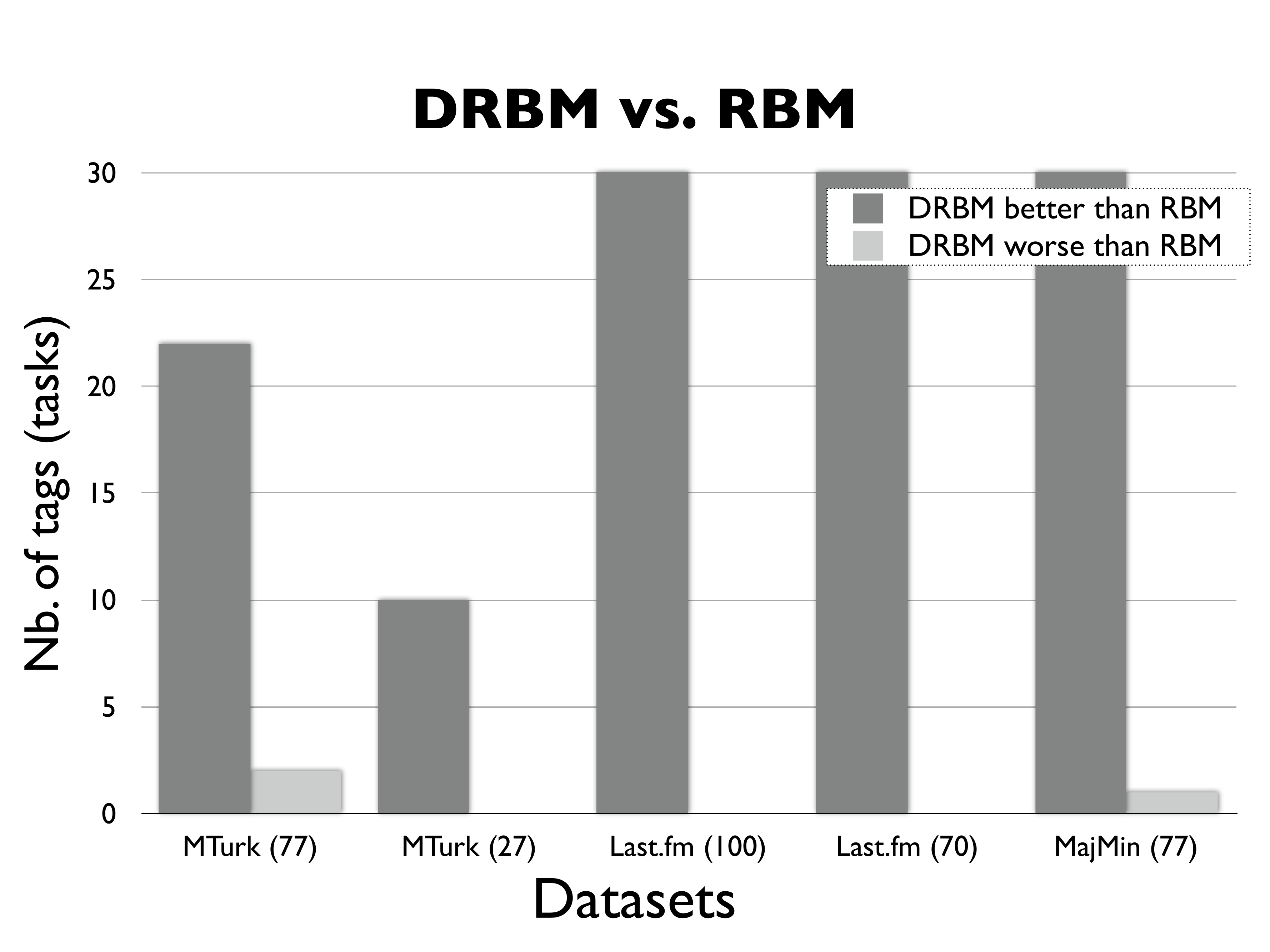} &
\includegraphics[height=\threeheight]{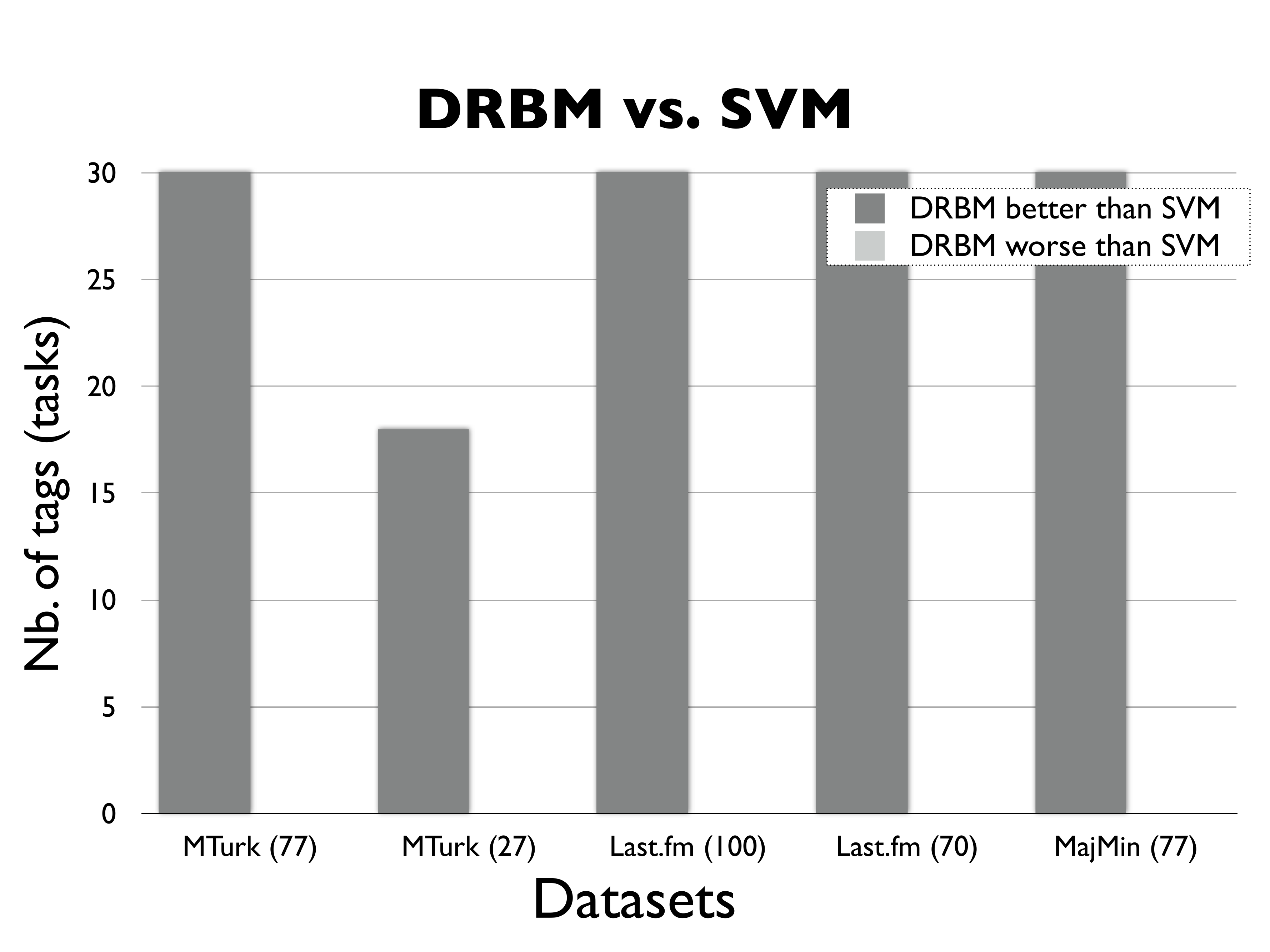} \\
\end{tabular}}
\end{center}
\caption{Comparison of discriminative restricted Boltzmann machine
  autotagging retrieval performance to multi-layered perceptron (MLP),
  logistic regression (LOG), (generative) restricted Boltzmann machine
  (RBM), and support vector machine (SVM).  Each bar shows performance
  on a different dataset and its height is the number of tags on which
  a two-sided paired t-test showed one algorithm to be significantly
  better than the other in terms of area under the ROC curve.
  Tags that were not significantly different are not included in
  this plot.} \label{fig:comparison_results}
\end{figure*}

\begin{figure*}[t]
\begin{center}
\begin{tabular}{ccc}
\includegraphics[height=\threeheight]{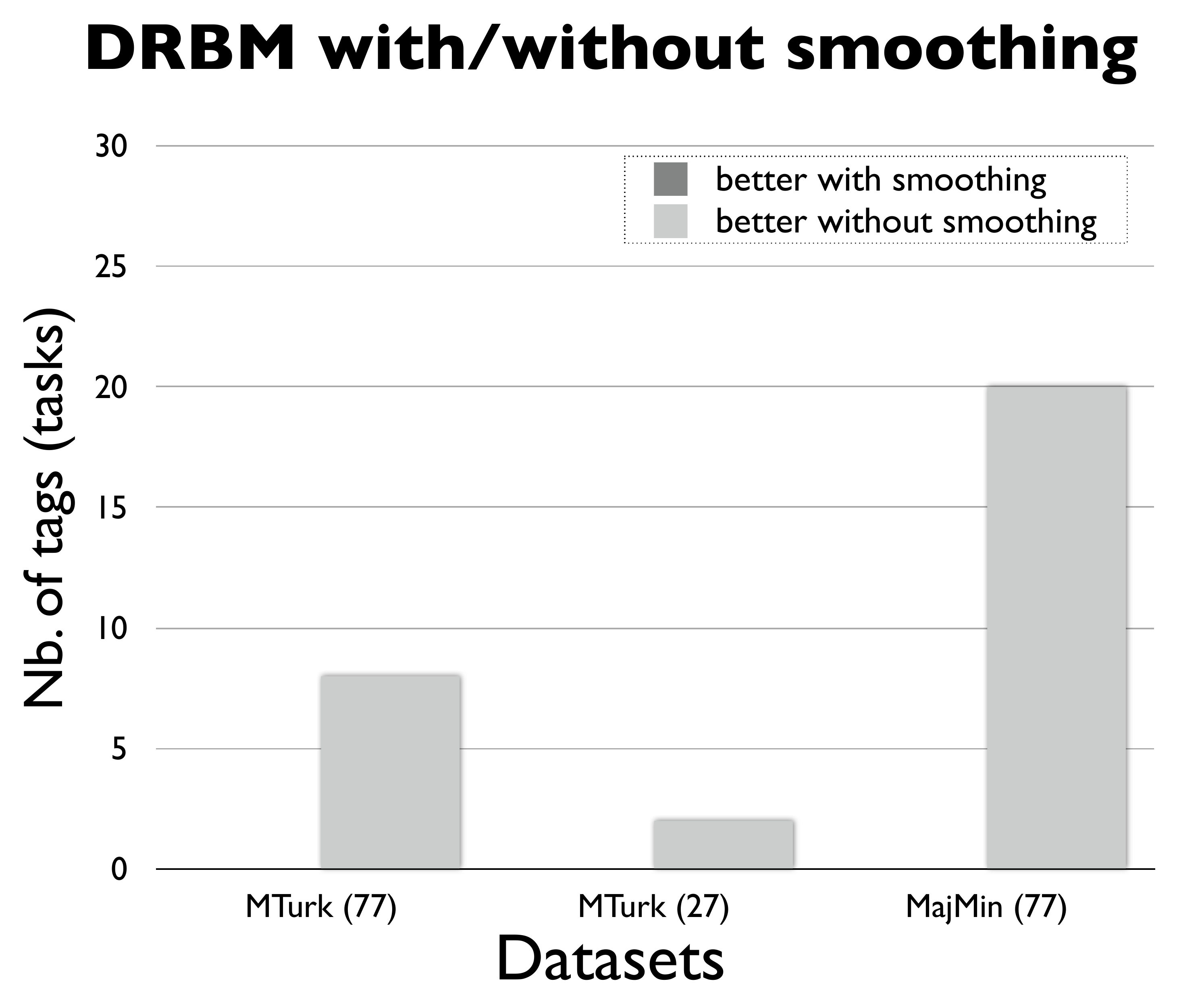} &
\includegraphics[height=\threeheight]{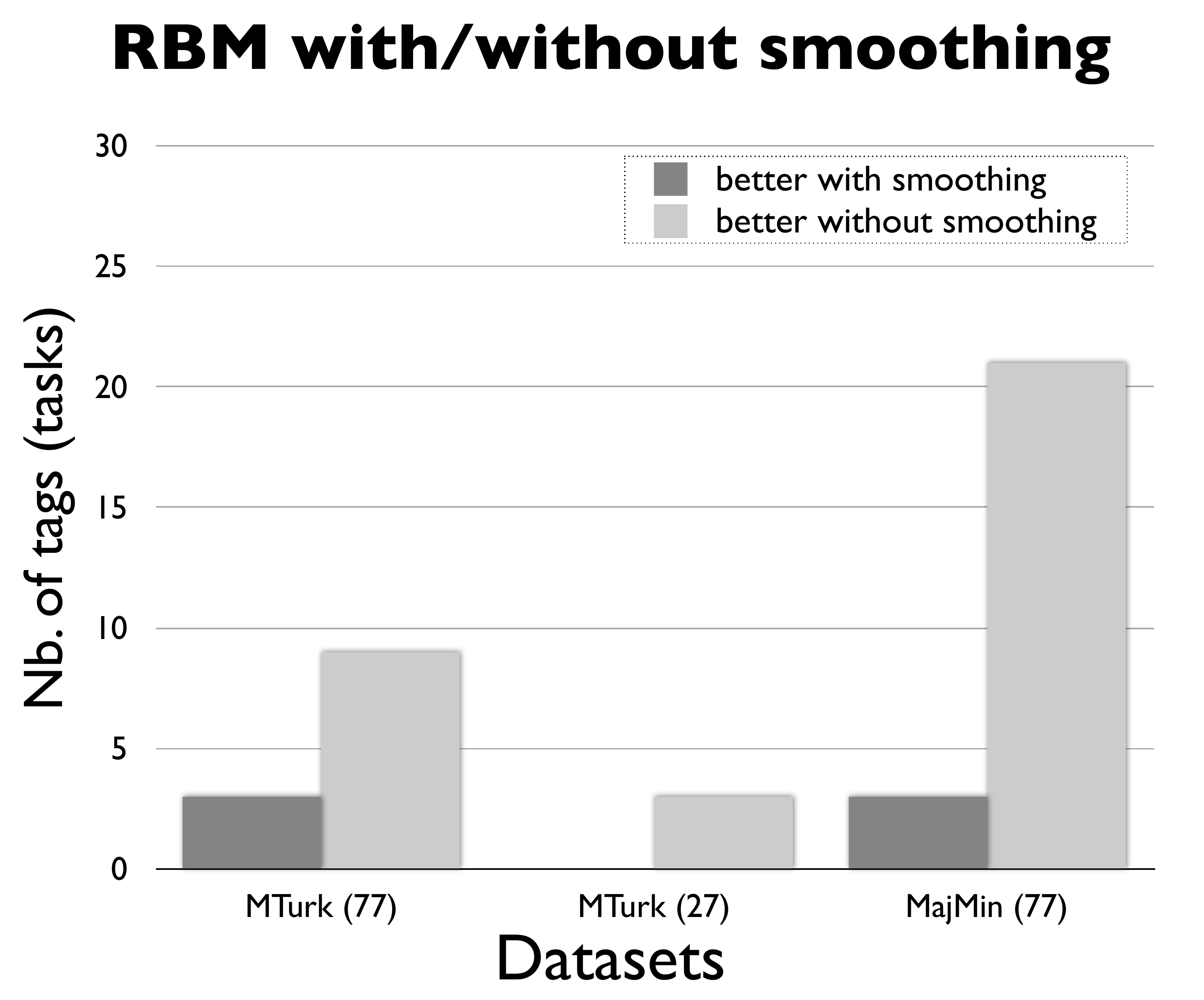} &
\includegraphics[height=\threeheight]{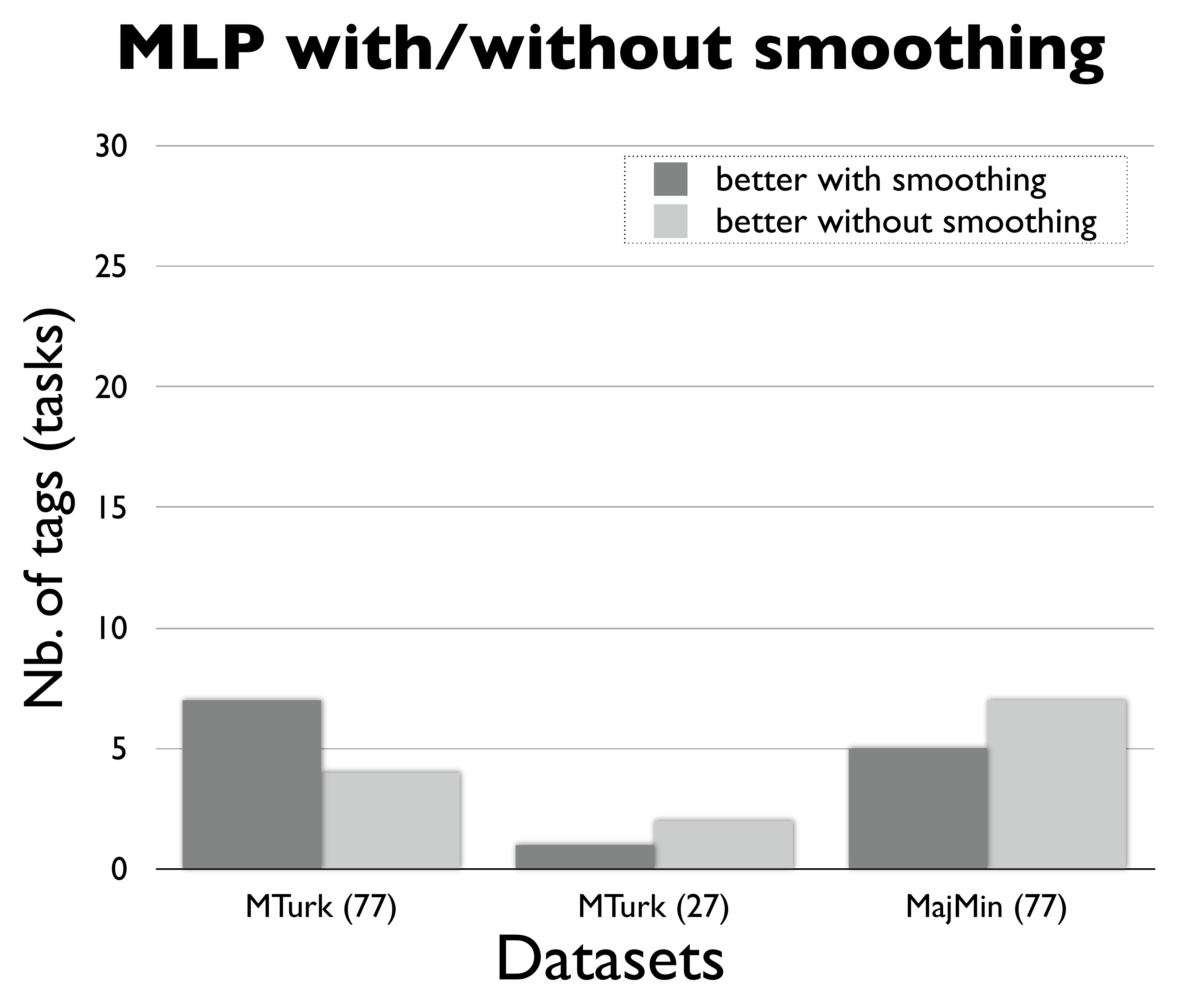} \\
\end{tabular}
\begin{tabular}{ccc}
\includegraphics[height=\threeheight]{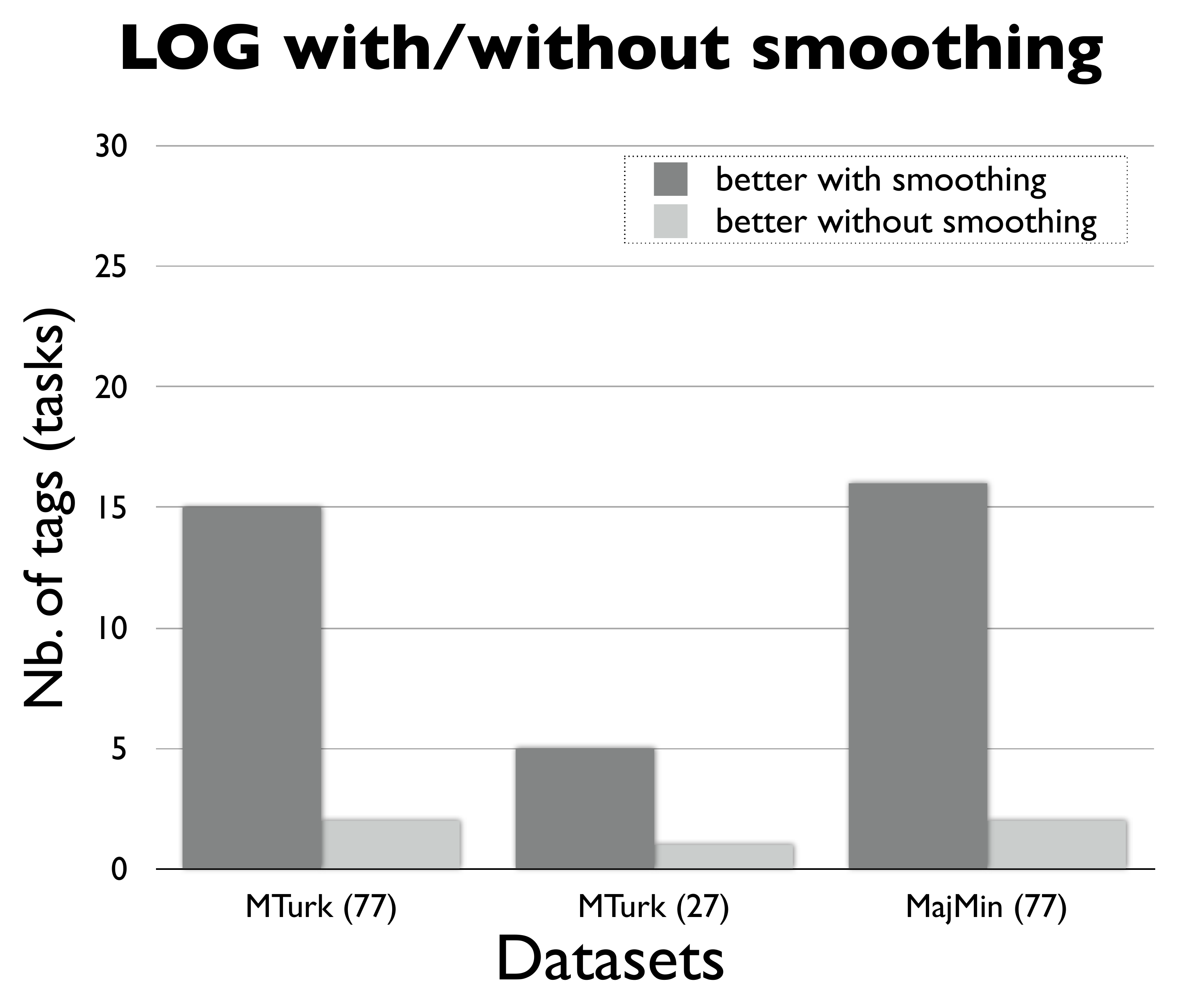} &
\includegraphics[height=\threeheight]{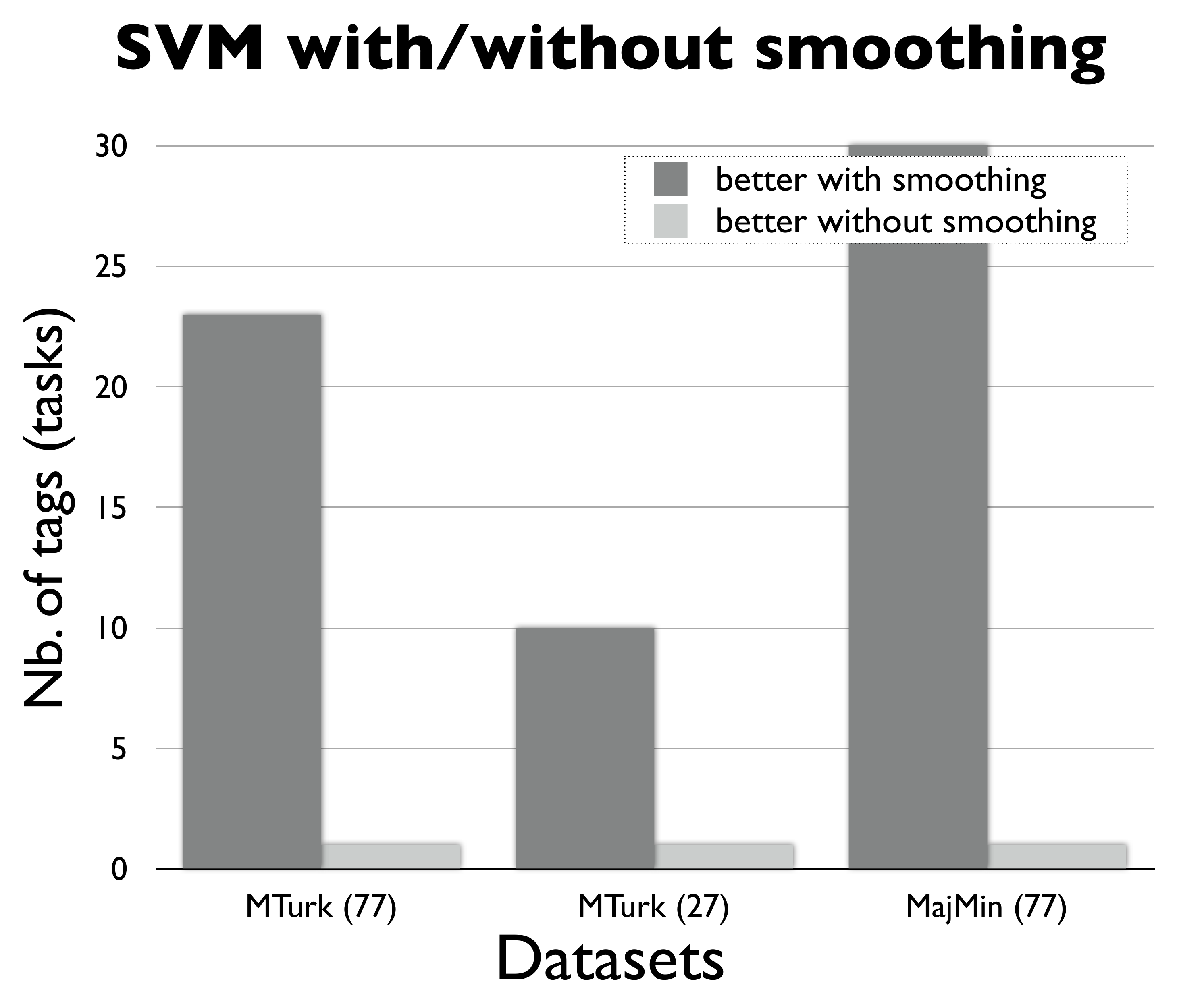} &
\\
\end{tabular}
\end{center}
\caption{Comparison of autotagging retrieval performance with and
  without conditional restricted Boltzmann machine-based smoothing for
  discriminative restricted Boltzmann machine (DRBM), multi-layer
  perceptron (MLP), (generative) restricted Boltzmann machine (RBM),
  logistic regression (LOG) and support vector machine (SVM).  Bars
  show number of tags on which one set of autotaggers was
  significantly better, as in
  Figure~\ref{fig:comparison_results}.} \label{fig:smoothing_results}
\end{figure*}

\begin{table}
\newcommand{\cs}{\hspace{0.8ex}}
\caption{Average AROC across tags as a percentage for each
  algorithm on each dataset with a specified number of tags (Tgs) and
  with and without tag smoothing (Sm).} \label{tab:smoothing_results}
\begin{center}
\begin{tabular}{l@{\hspace{.5ex}}c@{\hspace{.5ex}}c@{\cs}c@{\cs}c@{\cs}c@{\cs}c@{\cs}c}
\toprule
Dataset & Tgs & Sm & DRBM & MLP & RBM & LOG & SVM \\
\midrule
MTurk   & 27 & $-$        & 68.8 & 65.4 & 65.4 & 65.7 & 62.3 \\
MTurk   & 27 & $+$        & 68.4 & 65.6 & 64.6 & 66.7 & 66.0 \\
MTurk   & 77 & $-$        & 65.9 & 65.8 & 62.9 & 63.4 & 59.2 \\
MTurk   & 77 & $+$        & 65.9 & 66.1 & 62.4 & 64.6 & 64.0 \\
MajMin  & 77 & $-$        & 76.1 & 75.3 & 70.0 & 70.7 & 64.5 \\ 
MajMin  & 77 & $+$        & 74.8 & 74.8 & 68.2 & 73.3 & 71.5 \\
Last.fm & 70 & $-$        & 72.2 & 72.0 & 65.9 & 70.3 & 64.6 \\
Last.fm & 100 & $-$       & 72.4 & 72.4 & 66.1 & 70.2 & 64.5 \\
\bottomrule
\end{tabular}
\end{center}
\end{table}


\paragraph{Experiment 1} The first experiment measured the
effectiveness of different settings of the smoothing hyper-parameter in
loopy belief propagation, meant to aid the
convergence of the algorithm.
Figure~\ref{fig:params1} shows the mean area under the ROC curve
(AUC) on the MajorMiner dataset of discriminative RBMs trained
using loopy belief propagation (LBP) with different damping
factors. We use 10 training iterations. The plots
show that the damping factor does not change the accuracy of the 
model appreciably. Very similar results were obtained on the MTurk 
dataset (not shown), while for Last.fm
dataset we only use $\beta=0.9$ which performed best on MTurk.

\paragraph{Experiment 2} The second experiment compared discriminative
RBMs trained and tested with different combinations of approximations
to the intractable expectation in \eqref{eq:cond_grad}.
We use different approximations on train and test to fully explore the
space of possibilities.
The left plot in Figure~\ref{fig:params} shows the mean AUC of 
these discriminative RBMs on the MTurk dataset, while the right plot shows the
same results for MajorMiner. 
The four training approximations, in order of 
performance on MTurk, were contrastive divergence (CD), pseudo-likelihood (PL),
loopy belief propagation (LBP) and mean-field contrastive divergence
(MF). On MajorMiner the same order was preserved, except that loopy belief 
propagation outperformed pseudo-likelihood.
The testing approximations, in order of performance were
LBP and mean-field (MF).  The training approximation had a
larger impact on the final result than the testing approximation. For 
Last.fm we only used CD during training and LBP at test time.
We also found that the model is quite robust to the number of training or 
testing iterations for CD, MF or LBP.

\paragraph{Experiment 3} The third experiment compares the different
classifiers on the three datasets with and without tag smoothing.
We have also added a slight variation of the MTurk and Last.fm
datasets restricted to a subset 
of the most popular tags (27 for MTurk and 70 for Last.fm). 
Using a two-sided paired t-test
per tag, we compare all models to a discriminative restricted Boltzmann
machine trained on unsmoothed data.
The same test is done against all of the comparison models:
multi-layer perceptron (MLP), logistic regression (LOG), generative RBM
(RBM), and support vector machines (SVM). 
Figure~\ref{fig:comparison_results} shows the number of tags
on which the DRBM outperforms the other algorithm.  The DRBM
outperforms all of the other algorithms on many
more tags than it is outperformed.  The MLP is evenly matched to it on
the full Last.fm 100 dataset, but on the other four datasets, the DRBM
is significantly better on many more tags than it is worse.  The SVM
and logistic regression were previously the best performing
algorithms on these datasets.

Figure~\ref{fig:smoothing_results} shows the same analysis comparing
each classifier trained on the raw, user-supplied tags to the same classifier
trained on the tags smoothed by the proposed tag smoothing conditional
RBM.
Different subsets of the
auxiliary inputs were compared and the smoothing that gave the best
performance on the validation folds was selected.  Because of the size
of the Last.fm dataset, only the unsmoothed tags were tested. 
A number of interesting trends are visible in
Figure~\ref{fig:smoothing_results}.  First, the SVM and logistic
regression
models are helped by the tag smoothing.  This makes sense because
they treat each tag as a separate classification task and
cannot by themselves take advantage of the relationships between tags.
The MLP was sometimes helped by tag smoothing, but generally was not.
The fact that the RBMs were not helped by
the tag smoothing suggests that they are able to
capture by themselves the relationships between tags and do not need
the assistance of the tag smoothing.

\section{Conclusion}

This paper has described two applications of conditional restricted
Boltzmann machines to the task of autotagging music.  The
discriminative RBM was able to achieve a higher average area under the
ROC curve than the previously best known system for this problem, the
support vector machine, as well as the multi-layer perceptron and
logistic regression.  In order to be applied to this problem, the
discriminative RBM was generalized to the multi-label setting and
an in-depth analysis of four different learning algorithms for it were
evaluated.  The best results 
were obtained for a DRBM using contrastive divergence training and loopy
belief propagation at test time.  The performance
of the SVM was improved significantly, although not to the level of
the DRBM, by the purely textual tag smoothing conditional RBM.  Both
of these results demonstrate the power of modeling the relationships between
tags in autotagging systems.


\footnotesize
\bibliographystyle{unsrt}
\bibliography{aistats10}

\clearpage
\appendix

\section{Appendix: Pseudocode}

A discriminative RBM is based on the following energy function:
\begin{equation}
E(\y,\h,\x) = -\h^T U \y - \h^T W \x - b^T \y - c^T \h
\end{equation}
where $\x$ is conditioned on. From this energy function, we can define
a probability distribution over $\y$ and $\h$ as follows: $p(\y,\h \given \x) \propto e^{-E(\y,\h,\x)}$.

In the next sections, we describe the different approaches we evaluated
for training such an RBM.

\subsection{Contrastive Divergence}

The most straighforward approach is perhaps to train the RBM to maximise the
conditional log-likelihood of the associated target vector $\y$ by gradient
descent. To do so, we need to estimate the following gradient:
\begin{multline}
  \frac{\partial}{\partial \theta}\log p(\y_t \given \x_t) =
  -\expectation_{\h \given \y_t, \x_t} \left[
    \frac{\partial}{\partial \theta} E(\y_t,\h,\x_t) \right] \\
  + \expectation_{\y, \h \given \x_t}
  \left[\frac{\partial}{\partial \theta} E(\y,\h,\x_t)
  \right]. \label{eq:cond_grad-app}
\end{multline}
Since the second $\expectation_{\y, \h \given \x_t}$ is intractable,
we need to approximate it somehow. The contrastive divergence
algorithm~\cite{hinton02} proposes to replace this expectation by a
point estimate at a sample $\y^K$, obtained by running a Gibbs
sampling initialized at $\y$ for $K$ iterations. Given a sample $\y^K$
and given $\x_t$, the expectation with respect to $\h$ is now
tractable.

Algorithm~\ref{alg:cd} describes the associated training update,
given an example $(\y,\x)$. In
our notation, $a \leftarrow b$ means $a$ is set to value
$b$ and $a \sim p$ means $a$ is sampled from the distribution $p$.

\begin{algorithm}
\caption{Discriminative RBM training update using Contrastive Divergence. }
\label{alg:cd}
\begin{algorithmic}
  \STATE {\bfseries Input:} training pair $(\y,\x)$, number of iterations $K$ and learning rate $\lambda$
  \STATE  \# Positive phase
  \STATE $\y^0 \leftarrow \y$, $\widehat{\h}^0 \leftarrow \sigm(c +
  W \x + U \y^0)$ 
  \STATE
  \STATE  \# Negative phase ( we are doing CD-K here)
  \FOR{$K$ iterations}
  \STATE $\h^k \sim p(\h | \y^k, \x )$
  \STATE $\y^{k+1}  \sim p(\y | \h^k )$
  \STATE $\widehat{\h}^{k+1} \leftarrow \sigm(c + W \x + U \y^{k+1})$ 
  \ENDFOR
  \STATE
  \STATE \# Update
  \FOR{$\theta \in \Theta$}
  \STATE $\theta \leftarrow \theta - \lambda \left(
    \frac{\partial}{\partial \theta} E(\y^0,\x,\widehat{\h}^0) -
    \frac{\partial}{\partial \theta} E(\y^K,\x,\widehat{\h}^K)
  \right)$ 
  \ENDFOR
\end{algorithmic}
\end{algorithm}

\subsection{Mean-Field Contrastive Divergence}

A non-stochastic alternative to contrastive divergence is
mean-field contrastive divergence~\cite{WellingM2002}, where
samples are replaced by expectations. This procedure is detailed in Algoritm~\ref{alg:mfcd}.

\begin{algorithm}
\caption{Discriminative RBM training update using Mean-Field Contrastive Divergence. }
\label{alg:mfcd}
\begin{algorithmic}
  \STATE {\bfseries Input:} training pair $(\y,\x)$, number of iterations $K$ and learning rate $\lambda$
  \STATE  \# Positive phase
  \STATE $\widehat{\y}^0 \leftarrow \y$, $\widehat{\h}^0 \leftarrow \sigm(c +
  W \x + U \widehat{\y}^0)$ 
  \STATE
  \STATE  \# Negative phase ( we are doing MFCD-K here)
  \FOR{$K$ iterations}
  \STATE $\widehat{\y}^{k+1} \leftarrow \sigm(d + U^T \widehat{\h}^{k} )$ 
  \STATE $\widehat{\h}^{k+1} \leftarrow \sigm(c + W \x + U \widehat{\y}^{k+1})$ 
  \ENDFOR
  \STATE
  \STATE \# Update
  \FOR{$\theta \in \Theta$}
  \STATE $\theta \leftarrow \theta - \lambda \left(
    \frac{\partial}{\partial \theta} E(\widehat{y}^0,\x,\widehat{\h}^0) -
    \frac{\partial}{\partial \theta} E(\widehat{y}^K,\x,\widehat{\h}^K)
  \right)$ 
  \ENDFOR
\end{algorithmic}
\end{algorithm}

\subsection{Loopy Belief Propagation}

Instead of using a sample $\y^K$ to approximate the intractable
expectation, one could try to estimate directly the associated
marginals required by this expectation. Specifically, those marginals
are $p(y_j=1|\x)$, $p(h_k=1|\x)$ and $p(y_j=1,h_k=1|\x)$. Loopy belief
propagation~\cite{Murphy99loopybelief} is a popular algorithm for
approximating such marginals. Algorithm~\ref{alg:lbp} details this
procedure for the discriminative RBM. The given algorithm computes
messages in log-space and, for computational efficiency, messages are
normalized so that log-messages from zero-valued variables is 0 (hence
only messages from one-valued variables are passed).

\begin{algorithm*}
\caption{ Loopy Belief Propagation algorithm for inference in discriminative RBM}
\label{alg:lbp}
\begin{algorithmic}
  \STATE {\bfseries Input:} training pair $(\y,\x)$, number of iterations $K$ and damping factor $\beta$
  \STATE $m_{k j}^{\uparrow} \leftarrow 0,~~m_{k j}^{\downarrow} \leftarrow 0$~~~~$\forall~j,k$
  \STATE $c^{\rm data} \leftarrow c+W\x$
  \STATE 
  \STATE \# Update downwards (towards $\y$) and upwards (towards $\h$) messages
  \FOR{$K$ iterations}
  \STATE $m_{k j}^{\downarrow} \leftarrow \beta m_{k j}^{\downarrow} + (1-\beta) \log\left(1+(\exp(U_{k j})-1) ~\sigm(c^{\rm data}_k + \sum_{j^*\neq j} m_{k j^*}^{\uparrow})\right),$~~~~$\forall~j,k$
  \STATE $m_{k j}^{\uparrow} \leftarrow \beta m_{k j}^{\uparrow} + (1-\beta) \log\left(1+(\exp(U_{k j})-1) ~\sigm(d_j + \sum_{k^*\neq k} m_{k^* j}^{\downarrow})\right),$~~~~$\forall~j,k$
  \ENDFOR
  \STATE
  \STATE \# Compute estimated singleton and pair-wise marginals
  \STATE $p^{\rm LBP}(y_j=1|\x) \leftarrow \sigm(d_j + \sum_{k} m_{k j}^{\downarrow}),$~~~~$\forall~j$
  \STATE $p^{\rm LBP}(h_k=1|\x) \leftarrow \sigm(c^{\rm data}_k + \sum_{j} m_{k j}^{\uparrow}),$~~~~$\forall~k$
  \STATE 
  \STATE ${\rm num}^{01}_{k j} \leftarrow d_j + \sum_{k^*\neq k} m_{k^* j}^{\downarrow},~~{\rm num}^{10}_{k j} \leftarrow c^{\rm data}_k + \sum_{j^*\neq j} m_{k j^*}^{\uparrow},$~~~~$\forall~j,k$
  \STATE ${\rm num}^{11}_{k j} \leftarrow U_{k j} + {\rm num}^{10}_{k j} + {\rm num}^{01}_{k j},$~~~~$\forall~j,k$
  \STATE $p^{\rm LBP}(y_j=1,h_k=1|\x) = \exp({\rm num}^{11}_{k j})/(\exp({\rm num}^{11}_{k j}) + \exp({\rm num}^{01}_{k j}) + \exp({\rm num}^{10}_{k j})),$~~~~$\forall~j,k$
\end{algorithmic}
\end{algorithm*}

\subsection{Pseudo-likelihood training}

Finally, instead of approximately estimating the gradient of the log-likelihood
$\log p(\y_t \given \x_t)$, one could instead replace it by the pseudo-likelihood
objective~\cite{besag75}:
\begin{align}
\log PL(\y \given \x) &= \sum_j \log p(y_j \given
\y_{\backslash j}, \x) \\
= \sum_j &\log p(\y \given \x) - \log \left(
  p(\y \given \x) + p(\tilde{\y}_j \given \x) \right) \nonumber
\end{align}
and compute its gradient exactly. Algorithm~\ref{alg:pl} details the procedure
for updating the RBM according to that criteria.

\begin{algorithm*}
\caption{ Pseudo-likelihood training update algorithm in discriminative RBM}
\label{alg:pl}
\begin{algorithmic}
  \STATE {\bfseries Input:} training pair $(\y,\x)$, learning rate $\lambda$
  \STATE 
  \STATE \# Forward propagation
  \STATE $c^{\rm data} \leftarrow c+W\x + U\y$
  \STATE $\log PL(\y \given \x) \leftarrow 0$
  \FOR{$j$ from 1 to $|\y|$}
  \STATE $p(y_j \given \y_{\backslash j}, \x) \leftarrow
  \sigm \left\{ d_j + \sum_k \log \left[ 1+\exp(c^{\rm data}_k - U_{k j} y_j +
      U_{k j}) \right] -  \log \left[ 1+\exp(c^{\rm data}_k - U_{k j} y_j)
    \right] \right\}$
  \STATE $\log PL(\y \given \x) \leftarrow \log PL(\y \given \x) - y_j \log p(y_j \given \y_{\backslash j}, \x) - (1-y_j) \log (1-p(y_j \given \y_{\backslash j}, \x))$
  \ENDFOR
  \STATE
  \STATE \# Backward propagation
  \STATE $\partial U \leftarrow 0$, $\partial W \leftarrow 0$, $\partial c \leftarrow 0$
  \FOR{$j$ from 1 to $|\y|$}
  \STATE $\partial {\rm out}_j \leftarrow p(y_j \given \y_{\backslash j}, \x) - y_j,$ $\partial d_j \leftarrow \partial {\rm out}_j,$ $\partial {\rm hid} \leftarrow 0$
  \FOR{$k$ from 1 to $|\h|$}
  \STATE $\partial U_{k j} \leftarrow \partial U_{k j} + \partial {\rm out}_j ~((1-y_j)~\sigm(c^{\rm data}_k - U_{k j} y_j + U_{k j}) + y_j~ \sigm(c^{\rm data}_k - U_{k j} y_j))$
  \STATE $\partial {\rm hid}_k \leftarrow \partial {\rm out}_j ~(\sigm(c^{\rm data}_k - U_{k j} y_j + U_{k j}) - \sigm(c^{\rm data}_k - U_{k j} y_j))$
  \ENDFOR 
  \STATE $\partial U \leftarrow \partial U + \partial {\rm hid}~ \y^T,$ $\partial W \leftarrow \partial W + \partial {\rm hid}~ \x^T,$ $\partial c \leftarrow 
  \partial c + \partial {\rm hid} $
  \ENDFOR
  \STATE
  \STATE \# Update
  \STATE $U \leftarrow U - \lambda ~\partial U,$ $W \leftarrow W - \lambda ~\partial W,$ $c \leftarrow c - \lambda ~\partial c,$ $d \leftarrow d - \lambda ~\partial d$
\end{algorithmic}
\end{algorithm*}

\end{document}